\documentclass[11pt,a4paper]{article}

\usepackage{authblk}

\usepackage[hyperref]{acl2019}
\hypersetup{
    unicode=false,          
    pdftoolbar=true,        
    pdfmenubar=true,        
    pdffitwindow=false,     
    pdfstartview={FitH},    
    pdftitle={Enriching BERT with Knowledge Graph Embeddings for Document Classification}, 
    pdfauthor={Malte Ostendorff, Peter Bourgonje, Maria Berger, Julian Moreno-Schneider, Georg Rehm, Bela Gipp},     
    pdfsubject={Enriching BERT with Knowledge Graph Embeddings for Document Classification}, 
    pdfproducer={pdflatex}, 
    pdfkeywords={Document classification} {Text classification} {BERT} {Wikidata} {Knowledge Graph} {Embeddings}, 
    pdfnewwindow=true,      
    colorlinks=false,       
    linkcolor=red, 
    citecolor=green,        
    filecolor=magenta,      
    urlcolor=cyan           
}

\usepackage{times}
\usepackage{latexsym}
\usepackage{graphicx}
\usepackage{makecell}
\usepackage{booktabs}
\usepackage[utf8]{inputenc}
\usepackage[normalem]{ulem}
\useunder{\uline}{\ul}{}

\usepackage{authblk}

\usepackage{url}

\aclfinalcopy

\title{Enriching BERT with Knowledge Graph Embeddings for Document Classification}

\author[1,2]{\textbf{Malte Ostendorff}}
\author[1]{\textbf{Peter Bourgonje}}
\author[1]{\textbf{Maria Berger}}
\author[1]{\\ \textbf{Juli\'{a}n Moreno-Schneider}}
\author[1]{\textbf{Georg Rehm}}
\author[2]{\textbf{Bela Gipp}}

\affil[1]{Speech and Language Technology, DFKI GmbH, Germany}
\affil[ ]{\texttt{first.last@dfki.de}}

\affil[2]{University of Konstanz, Germany}
\affil[ ]{\texttt{first.last@uni-konstanz.de}}

\date{}

\begin{document}
\maketitle
\begin{abstract}

In this paper we focus on the classification of books using short descriptive texts (cover blurbs) and additional metadata.
Building upon BERT, a deep neural language model, we demonstrate how to combine text representations with metadata and knowledge graph embeddings, which encode author information.
Compared to the standard BERT approach we achieve considerably better results for the classification task.
For a more coarse-grained classification using eight labels we achieve an F1-score of 87.20, 
while a detailed classification using 343 labels yields an F1-score of 64.70. 
We make the source code and trained models of our experiments publicly available. 
\end{abstract}

\section{Introduction}
\label{sec:intro}

With ever-increasing amounts of data available, there is an increase in the need to offer tooling to speed up processing, and eventually making sense of this data. 
Because fully-automated tools to extract meaning from any given input to any desired level of detail have yet to be developed, this task is still at least supervised, and often (partially) resolved by humans; we refer to these humans as knowledge workers. 
Knowledge workers are professionals that have to go through large amounts of data and consolidate, prepare and process it on a daily basis. 
This data can originate from highly diverse portals and resources and depending on type or category, the data needs to be channelled through specific down-stream processing pipelines. 
We aim to create a platform for \textit{curation technologies} that can deal with such data from diverse sources and that provides natural language processing (NLP) pipelines tailored to particular content types and genres, rendering this initial classification an important sub-task.

In this paper, we work with the dataset of the 2019 GermEval shared task on hierarchical text classification~\cite{Remus2019} and use the predefined set of labels to evaluate our approach to this classification task\footnote{\url{https://www.inf.uni-hamburg.de/en/inst/ab/lt/resources/data/germeval-2019-hmc.html}}.

Deep neural language models have recently evolved to a successful method for representing text. 
In particular, Bidirectional Encoder Representations from Transformers (BERT; \citealp{Devlin2018}) outperformed previous state-of-the-art methods by a large margin on various NLP tasks.
We adopt BERT for text-based classification and extend the model with additional metadata provided in the context of the shared task, such as author, publisher, publishing date, etc.

A key contribution of this paper is the inclusion of additional (meta) data using a state-of-the-art approach for text processing. 
Being a transfer learning approach, it facilitates the task solution with external knowledge for a setup in which relatively little training data is available.
More precisely, we enrich BERT, as our pre-trained text representation model, with knowledge graph embeddings that are based on Wikidata \cite{vrandevcic2014wikidata}, add metadata provided by the shared task organisers (title, author(s), publishing date, etc.) and collect additional information on authors for this particular document classification task.
As we do not rely on text-based features alone but also utilize document metadata, we consider this as a document classification problem.
The proposed approach is an attempt to solve this problem exemplary for single dataset provided by the organisers of the shared task.

\section{Related Work}

A central challenge in work on genre classification is the definition of a both rigid (for theoretical purposes) and flexible (for practical purposes) mode of representation that is able to model various dimensions and characteristics of arbitrary text genres. 
The size of the challenge can be illustrated by the observation that there is no clear agreement among researchers regarding actual genre labels or their scope and consistency. 
There is a substantial amount of previous work on the definition of genre taxonomies, genre ontologies, or sets of labels \cite{biber1988,lee_bnc,sharoff18genres,Underwood2014,rehm2007}.
Since we work with the dataset provided by the organisers of the 2019 GermEval shared task, we adopt their hierarchy of labels as our genre palette. 
In the following, we focus on related work more relevant to our contribution.

With regard to \textbf{text and document classification}, BERT (Bidirectional Encoder Representations from Transformers) \cite{Devlin2018} is a pre-trained embedding model that yields state of the art results in a wide span of NLP tasks, such as question answering, textual entailment and natural language inference learning \cite{Artetxe2018}. 
\citet{Adhikari2019:DBLP:abs-1904-08398} are among the first to apply BERT to document classification. Acknowledging challenges like incorporating syntactic information, or predicting multiple labels, they describe how they adapt BERT for the document classification task. In general, they introduce a fully-connected layer over the final hidden state that contains one neuron each representing an input token, and further optimize the model choosing soft-max classifier parameters to weight the hidden state layer. They report state of the art results in experiments based on four popular datasets.
An approach exploiting Hierarchical Attention Networks is presented by~\citet{Yang2016}.  
Their model introduces a hierarchical structure to represent the hierarchical nature of a document. \citet{Yang2016} derive attention on the word and sentence level, which makes the attention mechanisms react flexibly to long and short distant context information during the building of the document representations.
They test their approach on six large scale text classification problems and outperform previous methods substantially by increasing accuracy by about 3 to 4 percentage points.
\citet{Aly2019} (the organisers of the GermEval 2019 shared task on hierarchical text classification) use shallow capsule networks, reporting that these work well on 
structured data for example in the field of visual inference, and outperform CNNs, LSTMs and SVMs in this area. 
They use the Web of Science (WOS) dataset and introduce a new real-world scenario dataset called Blurb Genre Collection (BGC)\footnote{Note that this is not the dataset used in the shared task.}. 

With regard to \textbf{external resources to enrich the classification task}, \citet{Zhang2019} experiment with external knowledge graphs to enrich embedding information in order to ultimately improve language understanding. They use structural knowledge represented by Wikidata entities and their relation to each other. A mix of large-scale textual corpora and knowledge graphs is used to further train language representation exploiting ERNIE \cite{sun19}, considering lexical, syntactic, and structural information. 
\citet{Wang2008UsingWK} propose and evaluate an approach to improve text classification with knowledge from Wikipedia. Based on a bag of words approach, they derive a thesaurus of concepts from Wikipedia and use it for document expansion. The resulting document representation improves the performance of an SVM classifier for predicting text categories.

\section{Dataset and Task}
\label{sec:data}

Our experiments are modelled on the GermEval 2019 shared task and deal with the classification of books. The dataset contains 20,784 German books. Each record has:
\begin{itemize}
\item A title.
\item A list of authors. The average number of authors per book is 1.13, with most books (14,970) having a single author and one outlier with 28 authors.
\item A short descriptive text (\textbf{blurb}) with an average length of 95 words.
\item A URL pointing to a page on the publisher's website.
\item An ISBN number.
\item The date of publication.
\end{itemize} 

The books are labeled according to the hierarchy used by the German publisher Random House. 
This taxonomy includes a mix of genre and topical categories. 
It has eight top-level genre categories, 93 on the second level and 242 on the most detailed third level.
The eight top-level labels are `Ganzheitliches Bewusstsein' (\textit{holistic awareness/consciousness}), `Künste' (\textit{arts}), `Sachbuch' (\textit{non-fiction}), `Kinderbuch \& Jugendbuch' (\textit{children and young adults}), `Ratgeber' (\textit{counselor/advisor}), `Literatur \& Unterhaltung' (\textit{literature and entertainment}), `Glaube \& Ethik' (\textit{faith and ethics}), `Architektur \& Garten' (\textit{architecture and garden}). 
We refer to the shared task description\footnote{\url{https://competitions.codalab.org/competitions/20139}} for details on the lower levels of the ontology.

Note that we do not have access to any of the full texts. 
Hence, we use the blurbs as input for BERT. 
Given the relatively short average length of the blurbs, this considerably decreases the amount of data points available for a single book.

The shared task is divided into two sub-task. 
Sub-task A is to classify a book, using the information provided as explained above, according to the top-level of the taxonomy, selecting one or more of the eight labels. 
Sub-task B is to classify a book according to the detailed taxonomy, specifying labels on the second and third level of the taxonomy as well (in total 343 labels).
This renders both sub-tasks a multi-label classification task.

\section{Experiments}
As indicated in Section~\ref{sec:intro}, we base our experiments on BERT in order to explore if it can be successfully adopted to the task of book or document classification. 
We use the pre-trained models and enrich them with additional metadata and tune the models for both classification sub-tasks.

\subsection{Metadata Features}
\label{sec:metadata_features}

In addition to the metadata provided by the organisers of the shared task (see Section~\ref{sec:data}), we add the following features.

\begin{itemize}
\item Number of authors.
\item Academic title (Dr.~or Prof.), if found in author names (0 or 1).
\item Number of words in title.
\item Number of words in blurb.
\item Length of longest word in blurb.
\item Mean word length in blurb.
\item Median word length in blurb.
\item Age in years after publication date.
\item Probability of first author being male or female based on the \textit{Gender-by-Name} dataset\footnote{Probability of given names being male/female based on US names from 1930-2015. \url{https://data.world/howarder/gender-by-name}}. Available for 87\% of books in training set (see Table~\ref{tab:data_availability}).
\end{itemize} 

The statistics (length, average, etc.) regarding blurbs and titles are added in an attempt to make certain characteristics explicit to the classifier. For example, books labeled `Kinderbuch \& Jugendbuch' (\textit{children and young adults}) have a title that is on average 5.47 words long, whereas books labeled `Künste' (\textit{arts}) on average have shorter titles of 3.46 words.
The binary feature for academic title is based on the assumption that academics are more likely to write non-fiction.
The gender feature is included to explore (and potentially exploit) whether or not there is a gender-bias for particular genres.

\begin{table}[t]
\begin{tabular}{|l|c|c|c|}
\hline
    & \multicolumn{1}{c|}{\textbf{Train}} & \multicolumn{1}{c|}{\textbf{Validation}} & \multicolumn{1}{c|}{\textbf{Test}} \\ \hline
\textbf{Gender}            
    & \makecell[c]{12,681 \\  (87\%) }                    
    & \makecell[c]{1,834 \\(88\%)}                              
    & \makecell[c]{3,641 \\ (88\%)}                        
    \\ \hline
\textbf{Author emb.} 
    & \makecell[c]{10,407 \\ (72\%)}                        
    & \makecell[c]{1,549 \\ (75\%)}                           
    & \makecell[c]{3,010 \\ (72\%)}                        \\ \hline
\textbf{Total books}            & 14,548                               & 2,079                                     & 4,157 \\ \hline
\end{tabular}
\caption{Availability of additional data with respect to the dataset (relative numbers in parenthesis).}\label{tab:data_availability}
\end{table}

\subsection{Author Embeddings}
\label{sec:author_embeddings}

\begin{figure}[t]
\centering
\includegraphics[width=0.49\textwidth,trim={20cm 3cm 20cm 3cm},clip]{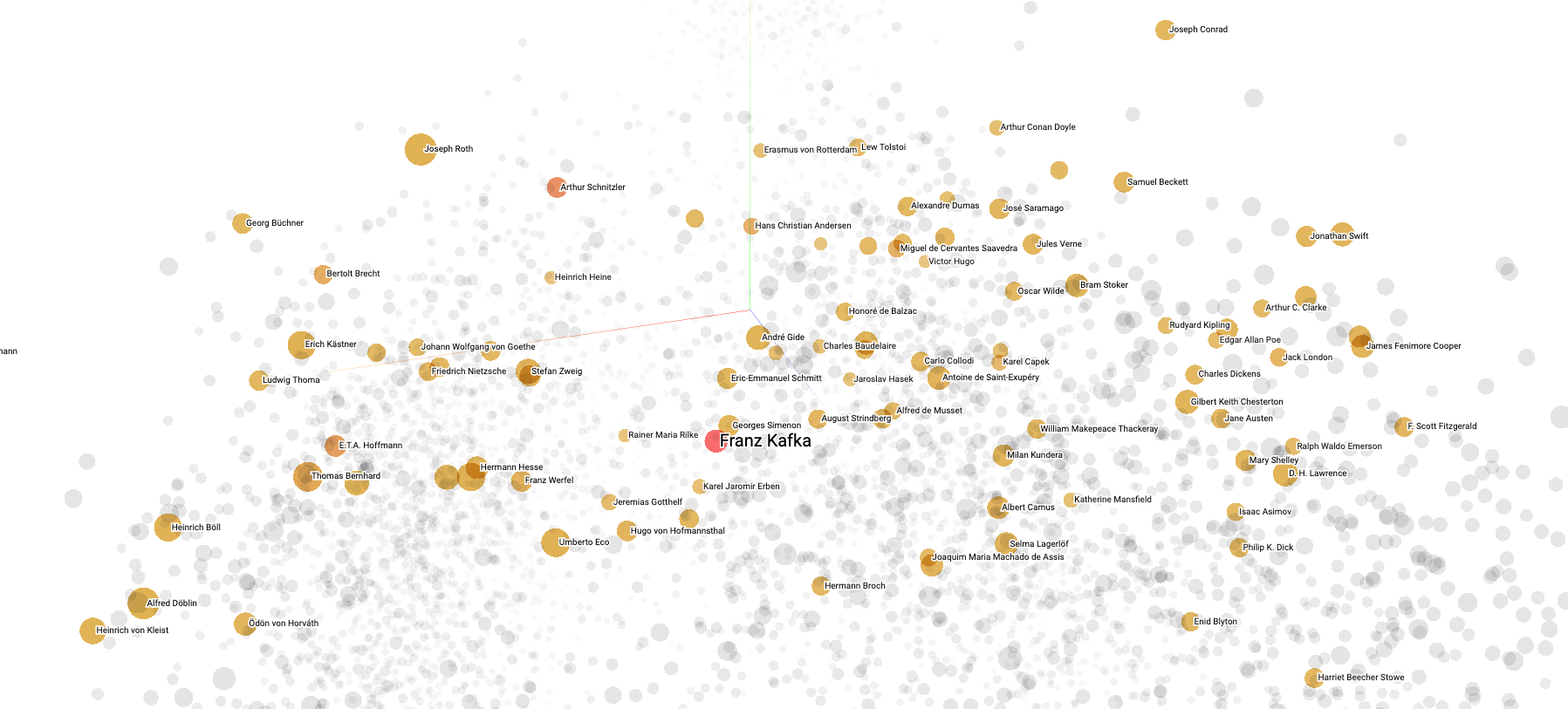}
\caption{\label{fig:author_embeddings}Visualization of Wikidata embeddings for \textit{Franz Kafka} (3D-projection with PCA)\footnote{\textit{link omitted for anonymity}}. Nearest neighbours in original 200D space: \textit{Arthur Schnitzler}, \textit{E.T.A Hoffmann} and \textit{Hans Christian Andersen}. }
\end{figure}

Whereas one should not judge a book by its cover, we argue that additional information on the author can support the classification task. 
Authors often adhere to their specific style of writing and are likely to specialize in a specific genre.

To be precise, we want to include author identity information, which can be retrieved by selecting particular properties from, for example, the Wikidata knowledge graph (such as date of birth, nationality, or other biographical features). 
A drawback of this approach, however, is that one has to manually select and filter those properties that improve classification performance. This is why, instead, we follow a more generic approach and utilize automatically generated graph embeddings as author representations.

Graph embedding methods create dense vector representations for each node such that distances between these vectors predict the occurrence of edges in the graph. The node distance can be interpreted as topical similarity between the corresponding authors.

We rely on pre-trained embeddings based on PyTorch BigGraph~\cite{Lerer2019}. 
The graph model is trained on the full Wikidata graph, using a translation operator to represent relations\footnote{Pre-trained Knowledge Graph Embeddings. \url{https://github.com/facebookresearch/PyTorch-BigGraph\#pre-trained-embeddings}}. 
Figure~\ref{fig:author_embeddings} visualizes the locality of the author embeddings.

To derive the author embeddings, we look up Wikipedia articles that match with the author names and map the articles to the corresponding Wikidata items\footnote{Mapping Wikipedia pages to Wikidata IDs and vice versa. \url{https://github.com/jcklie/wikimapper}}. 
If a book has multiple authors, the embedding of the first author for which an embedding is available is used.
Following this method, we are able to retrieve embeddings for 72\% of the books in the training and test set (see Table~\ref{tab:data_availability}).

\subsection{Pre-trained German Language Model}

Although the pre-trained BERT language models are multilingual and, therefore, support German, we rely on a BERT model that was exclusively pre-trained on German text, as published by the German company Deepset AI\footnote{Details on BERT-German training procedure: \url{https://deepset.ai/german-bert}}. 
This model was trained from scratch on the German Wikipedia, news articles and court decisions\footnote{German legal documents used to train BERT-German: \url{http://openlegaldata.io/research/2019/02/19/court-decision-dataset.html}}. 
Deepset AI reports better performance for the German BERT models compared to the multilingual models on previous German shared tasks (GermEval2018-Fine and GermEval 2014). 

\subsection{Model Architecture}
\label{sec:model}

\begin{figure}[t]
\centering
\includegraphics[width=0.5\textwidth,trim={1cm 2cm 1cm 1.5cm},clip]{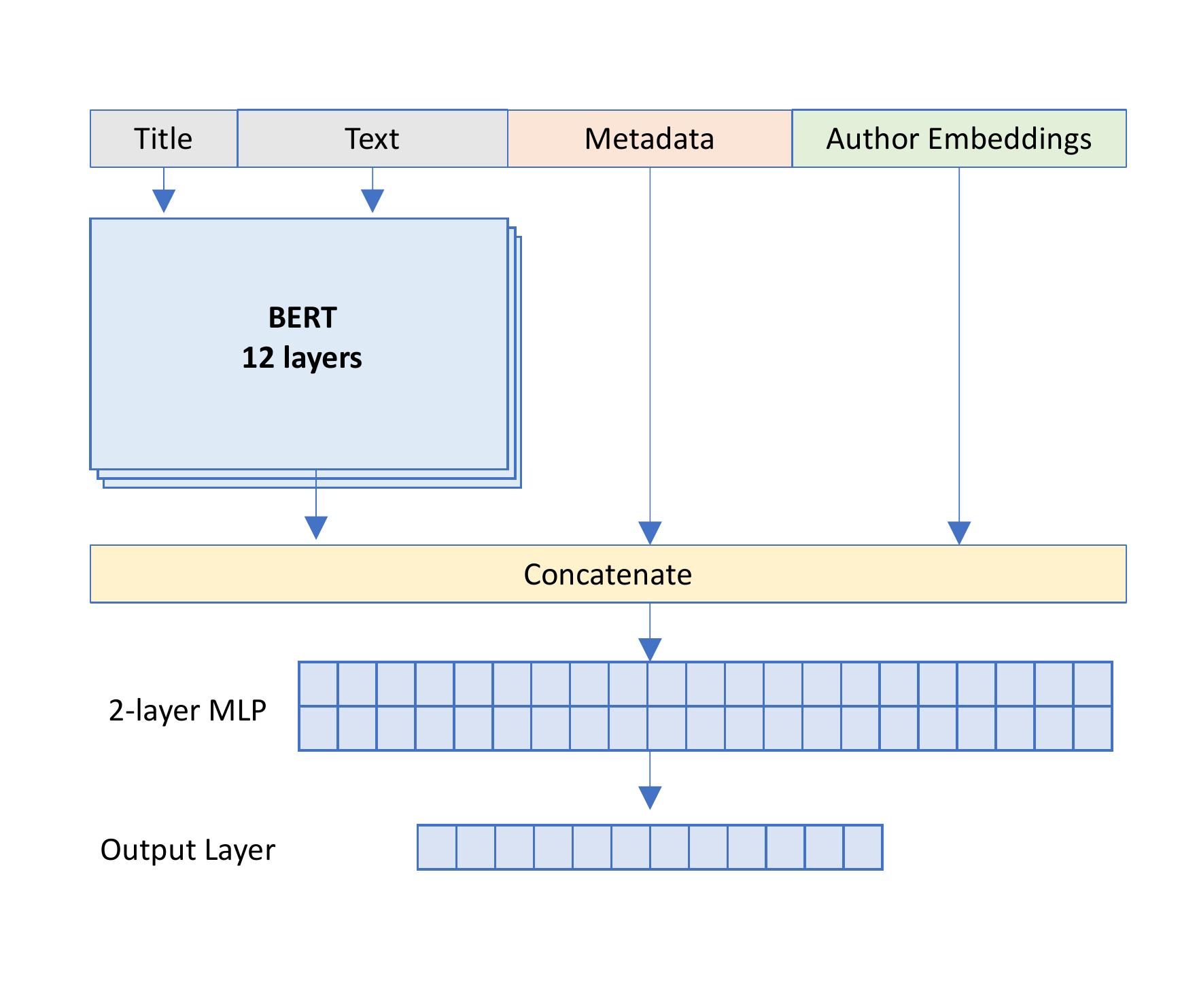}
\caption{\label{fig:architecture}Model architecture used in our experiments. Text-features are fed through BERT, concatenated with metadata and author embeddings and combined in a multilayer perceptron (MLP).}
\end{figure}

Our neural network architecture, shown in Figure~\ref{fig:architecture}, resembles the original BERT model \cite{Devlin2018} and combines text- and non-text features with a multilayer perceptron (MLP). 

The BERT architecture uses 12 hidden layers, each layer consists of 768 units. To derive contextualized representations from textual features, the book title and blurb are concatenated and then fed through BERT. To minimize the GPU memory consumption, we limit the input length to 300 tokens (which is shorter than BERT's hard-coded limit of 512 tokens). 
Only 0.25\% of blurbs in the training set consist of more than 300 words, so this cut-off can be expected to have minor impact.

The non-text features are generated in a separate preprocessing step. 
The metadata features are represented as a ten-dimensional vector (two dimensions for gender, see Section~\ref{sec:metadata_features}). 
Author embedding vectors have a length of 200 (see Section~\ref{sec:author_embeddings}).
In the next step, all three representations are concatenated and passed into a MLP with two layers, 1024 units each and ReLu activation function. 
During training, the MLP is supposed to learn a non-linear combination of its input representations.
Finally, the output layer does the actual classification.
In the SoftMax output layer each unit corresponds to a class label. 
For sub-task A the output dimension is eight.
We treat sub-task B as a standard multi-label classification problem, i.\,e., we neglect any hierarchical information. 
Accordingly, the output layer for sub-task B has 343 units.
When the value of an output unit is above a given threshold the corresponding label is predicted, whereby thresholds are defined separately for each class.
The optimum was found by varying the threshold in steps of $0.1$ in the interval from $0$ to $1$.

\subsection{Implementation}
\label{sec:implementation}

Training is performed with batch size $b=16$, dropout probability $d=0.1$, learning rate $\eta=2^{-5}$ (Adam optimizer) and 5 training epochs. These hyperparameters are the ones proposed by \citet{Devlin2018} for BERT fine-tuning. We did not experiment with hyperparameter tuning ourselves except for optimizing the classification threshold for each class separately. 
All experiments are run on a GeForce GTX 1080 Ti (11 GB), whereby a single training epoch takes up to 10min. If there is no single label for which prediction probability is above the classification threshold, the most popular label (\textit{Literatur \& Unterhaltung}) is used as prediction.

\subsection{Baseline} 
\label{sec:baseline}

To compare against a relatively simple baseline, we implemented a Logistic Regression classifier chain from scikit-learn \cite{scikit-learn}.
This baseline uses the text only and converts it to TF-IDF vectors. As with the BERT model, it performs 8-class multi-label classification for sub-task A and 343-class multi-label classification for sub-task B, ignoring the hierarchical aspect in the labels.

\section{Results}
\label{sec:results}

\begin{table*}[t]
\centering

\setlength\tabcolsep{10px}
\resizebox{\textwidth}{!}{
\begin{tabular}{|l|c|c|c|c|c|c|}
\hline
 & \multicolumn{3}{c|}{\makecell[c]{Sub-Task A -- \textit{8 labels}}}                                                                          & \multicolumn{3}{c|}{\makecell[c]{Sub-Task B -- \textit{343 labels}}}                                                                          \\ \hline
 \textbf{Model / Features}
 & \multicolumn{1}{c|}{\textbf{F1}} & \multicolumn{1}{c|}{\textbf{Prec.}} & \multicolumn{1}{c|}{\textbf{Recall}} & \multicolumn{1}{c|}{\textbf{F1}} & \multicolumn{1}{c|}{\textbf{Prec.}} & \multicolumn{1}{c|}{\textbf{Recall}} \\ \hline
(1) BERT-German + Metadata + Author
& {\ul 87.20}                      
& 88.76                           
& {\ul 85.70}                     
& {\ul 64.70}                      
& 83.78                           
& 52.70                           
\\ \hline

(2) BERT-German + Metadata
& 86.90                            
& {\ul 89.65}                     
& 84.30                           
& 63.96                            
& {\ul 83.94}                     
& 51.67                           
\\ \hline

(3) BERT-German + Author     
& 86.84                            
& 89.02                           
& 84.75                           
& 64.41                            
& 82.02                           
& {\ul 53.03}                     
\\ \hline

(4) BERT-German         
& 86.65                            
& 89.65                           
& 83.86                           
& 60.51                            
& 83.44                           
& 47.47                           
\\ \hline

(5) BERT-Base-Multilingual-Cased
& 83.94                            
& 86.31                           
& 81.70                           
& 54.08                            
& 82.63                           
& 40.19                           
\\ \hline

(6) Author                   
& 61.99                            
& 75.59                           
& 52.54                           
& 32.13                            
& 72.39                           
& 20.65                           
\\ \hline

(7) Baseline
& 77.00                            
& 79.00                           
& 74.00                           
& 45.00                            
& 67.00                           
& 34.00          

\\ \hline
\hline

Results of best model (1) on test set
& 88.00                            
& 85.00                           
& 86.00                           
& 78.00                            
& 52.00                           
& 62.00                          
\\ \hline

\end{tabular}}

\caption{Evaluation scores (micro avg.) on validation set with respect to the features used for classification. The model with BERT-German, metadata and author embeddings yields the highest F1-scores on both tasks and was accordingly submitted to the GermEval 2019 competition. The scores in the last row are the result on the test set as reported by \citealp{Remus2019}.}\label{tab:results}

\end{table*}

Table~\ref{tab:results} shows the results of our experiments. As prescribed by the shared task, the essential evaluation metric is the micro-averaged F1-score. All scores reported in this paper are obtained using models that are trained on the training set and evaluated on the validation set. For the final submission to the shared task competition, the best-scoring setup is used and trained on the training and validation sets combined.

We are able to demonstrate that incorporating metadata features and author embeddings leads to better results for both sub-tasks. 
With an F1-score of 87.20 for task A and 64.70 for task B, the setup using BERT-German with metadata features and author embeddings (1) outperforms all other setups. 
Looking at the precision score only, BERT-German with metadata features (2) but without author embeddings performs best. 

In comparison to the baseline (7), our evaluation shows that deep transformer models like BERT considerably outperform the classical TF-IDF approach, also when the input is the same (using the title\footnote{The baseline model uses the blurbs only, without the title, but we do not expect that including the title in the input would make up for the considerable gap between the two.} and blurb only). BERT-German (4) and BERT-Multilingual (5) are only using text-based features (title and blurb), whereby the text representations of the BERT-layers are directly fed into the classification layer.

To establish the information gain of author embeddings, we train a linear classifier on author embeddings, using this as the only feature. 
The author-only model (6) is exclusively evaluated on books for which author embeddings are available, so the numbers are based on a slightly smaller validation set. With an F1-score of 61.99 and 32.13 for sub-tasks A and B, respectively, the author model yields the worst result. However, the information contained in the author embeddings help improve performance, as the results of the best-performing setup show.
When evaluating the best model (1) only on books for that author embeddings are available, 
we find a further improvement with respect to F1 score (task A: from 87.20 to 87.81; task B: 64.70 to 65.74).

\section{Discussion}

\begin{table*}
\small
\centering
\begin{tabular}{p{0.22\linewidth}p{0.35\linewidth}p{0.34\linewidth}} \toprule\addlinespace
\multicolumn{1}{c}{\textbf{Title / Author}}                   
& \multicolumn{1}{c}{\textbf{Correct Labels}}                                                                                                                                                                                          & \multicolumn{1}{c}{\textbf{Predicted Labels}}                            
\\ \addlinespace\midrule\addlinespace

\makecell[l]{\textit{Coenzym Q10}  \\ Dr. med. Gisela Rauch-Petz }
& \makecell[l]{ Ratgeber (I); Gesundheit \& Ern\"ahrung~(II) }                                                                                                                                                                
& \makecell[l]{ Gesundheit \& Ern\"ahrung~(II) }
\\ \addlinespace

\makecell[l]{\textit{Gelebte Wertsch\"atzung}        \\ Barbara von Meibom  } 
& \makecell[l]{Glaube \& Ethik~(I); \\ Psychologie \& Spiritualit\"at (II) }                                                                                                                                                          & \makecell[l]{ Sachbuch (I); Politik \& Gesellschaft (II) }
\\ \addlinespace

\makecell[l]{\textit{Wie Romane entstehen} \\Hanns-Josef Ortheil, \\ Klaus Siblewski } 
& \makecell[l]{Literatur \& Unterhaltung~(I); Sachbuch (I); \\ Romane \& Erz\"ahlungen~(II); \\ Briefe, Essays, Gespr\"ache~(II) }                                                                                     & \makecell[l]{Literatur \& Unterhaltung~(I)  }
\\ \addlinespace
\makecell[l]{\textit{Das Grab ist erst der Anfang} \\  Kathy Reichs }  
& \makecell[l]{Literatur \& Unterhaltung~(I); \\ Krimi \& Thriller~(II)      }                                                                                                                                                      & \makecell[l]{ Literatur \& Unterhaltung~(I); \\ Krimi \& Thriller~(II)} \\ \addlinespace\bottomrule
\end{tabular}

\caption{Book examples and their correct and predicted labels. Hierarchical label level is in parenthesis.}\label{tab:example}
\end{table*}

\begin{figure}[ht]
\centering
\includegraphics[width=0.5\textwidth,trim={0 0 1cm 0},clip]{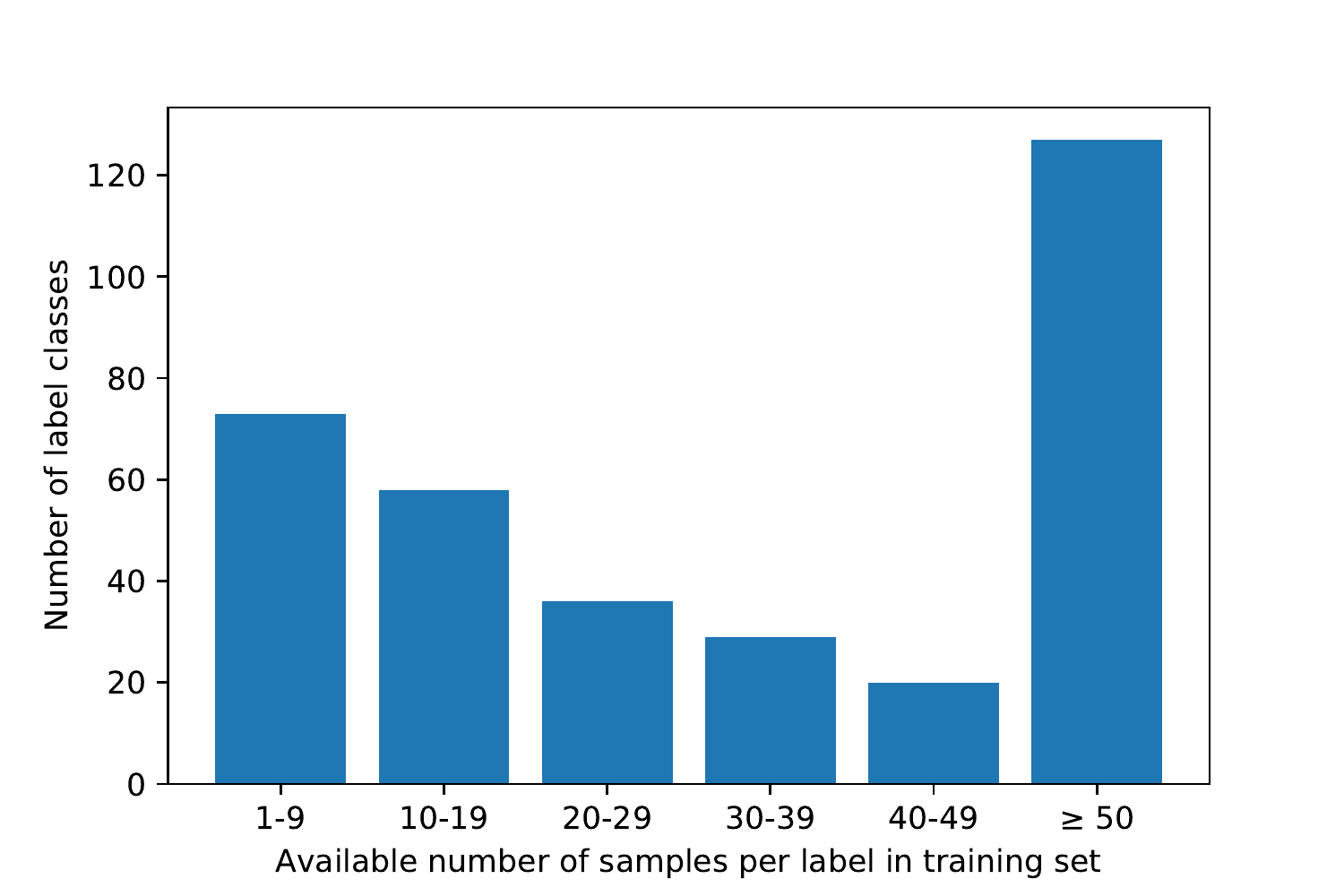}
\caption{\label{fig:samples_per_label}In sub-task B for many low-hierarchical labels only a small number of training samples exist, making it more difficult to predict the correct label. }
\end{figure}

The best performing setup uses BERT-German with metadata features and author embeddings. In this setup the most data is made available to the model, indicating that, perhaps not surprisingly, more data leads to better classification performance. We expect that having access to the actual text of the book will further increase performance. 
The average number of words per blurb is 95 and only 0.25\% of books exceed our cut-off point of 300 words per blurb. 
In addition, the distribution of labeled books is imbalanced, i.e. for many classes only a single digit number of training instances exist (Fig.~\ref{fig:samples_per_label}). 
Thus, this task can be considered a low resource scenario, where including related data (such as author embeddings and author identity features such as gender and academic title) or making certain characteristics more explicit (title and blurb length statistics) helps. Furthermore, it should be noted that the blurbs do not provide summary-like abstracts of the book, but instead act as teasers, intended to persuade the reader to buy the book. 

As reflected by the recent popularity of deep transformer models, they considerably outperform the Logistic Regression baseline using TF-IDF representation of the blurbs. However, for the simpler sub-task A, the performance difference between the baseline model and the multilingual BERT model is only six points, while consuming only a fraction of BERT's computing resources. The BERT model trained for German (from scratch) outperforms the multilingual BERT model by under three points for sub-task A and over six points for sub-task B, confirming the findings reported by the creators of the BERT-German models for earlier GermEval shared tasks.

While generally on par for sub-task A\footnote{Except for the Author-only setup.}, for sub-task B there is a relatively large discrepancy between precision and recall scores. In all setups, precision is considerably higher than recall. We expect this to be down to the fact that for some of the 343 labels in sub-task B, there are very few instances. This means that if the classifier predicts a certain label, it is likely to be correct (i.\,e., high precision), but for many instances having low-frequency labels, this low-frequency label is never predicted (i.\,e., low recall).

As mentioned in Section~\ref{sec:model}, we neglect the hierarchical nature of the labels and flatten the hierarchy (with a depth of three levels) to a single set of 343 labels for sub-task B. We expect this to have negative impact on performance, because it allows a scenario in which, for a particular book, we predict a label from the first level and also a non-matching label from the second level of the hierarchy. The example \textit{Coenzym Q10} (Table~\ref{tab:example}) demonstrates this issue. While the model correctly predicts the second level label \textit{Gesundheit \& Ern\"ahrung} (health \& diet), it misses the corresponding first level label \textit{Ratgeber} (advisor). Given the model's tendency to higher precision rather than recall in sub-task B, as a post-processing step we may want to take the most detailed label (on the third level of the hierarchy) to be correct and manually fix the higher level labels accordingly. We leave this for future work and note that we expect this to improve performance, but it is hard to say by how much. We hypothesize that an MLP with more and bigger layers could improve the classification performance. However, this would increase the number of parameters to be trained, and thus requires more training data (such as the book's text itself, or a summary of it).

\section{Conclusions and Future Work}

In this paper we presented a way of enriching BERT with knowledge graph embeddings and additional metadata. 
Exploiting the linked knowledge that underlies Wikidata improves performance for our task of document classification.
With this approach we improve the standard BERT models by up to four percentage points in accuracy. 
Furthermore, our results reveal that with task-specific information such as author names and publication metadata improves the classification task essentially compared a text-only approach.
Especially, when metadata feature engineering is less trivial, adding additional task-specific information from an external knowledge source such as Wikidata can help significantly.
The source code of our experiments and the trained models are publicly available\footnote{\url{https://ostendorff.org/r/germeval19}}.

Future work comprises the use of hierarchical information in a post-processing step to refine the classification. Another promising approach to tackle the low resource problem for task B would be to use label embeddings. Many labels are similar and semantically related. The relationships between labels can be utilized to model in a joint embedding space \cite{Augenstein2018}. However, a severe challenge with regard to setting up label embeddings is the quite heterogeneous category system that can often be found in use online. The Random House taxonomy (see above) includes category names, i.\,e., labels, that relate to several different dimensions including, among others, genre, topic and function.

This work is done in the context of a larger project that develops a platform for curation technologies. Under the umbrella of this project, the classification of pieces of incoming text content according to an ontology is an important step that allows the routing of this content to particular, specialized processing workflows, including parameterising the included pipelines. 
Depending on content type and genre, it may make sense to apply OCR post-processing (for digitized books from centuries ago), machine translation (for content in languages unknown to the user), information extraction, or other particular and specialized procedures. 
Constructing such a generic ontology for digital content is a challenging task, and classification performance is heavily dependent on input data (both in shape and amount) and on the nature of the ontology to be used (in the case of this paper, the one predefined by the shared task organisers).
In the context of our project, we continue to work towards a maximally generic content ontology, and at the same time towards applied classification architectures such as the one presented in this paper. 

\section*{Acknowledgments}
This research is funded by the German Federal Ministry of Education and Research (BMBF) through the ``Unternehmen Region'', instrument ``Wachstumskern'' QURATOR (grant no. 03WKDA1A). We would like to thank the anonymous reviewers for comments on an earlier version of this manuscript.


\bibliography{acl2019}
\bibliographystyle{acl_natbib}

\end{document}